\documentclass[conference]{IEEEtran}
\IEEEoverridecommandlockouts

\usepackage{cite}
\usepackage{amsmath,amssymb,amsfonts}
\usepackage{algorithmic}
\usepackage{graphicx}
\usepackage{textcomp}
\usepackage{xcolor}
\usepackage{float}
\usepackage{authblk}
\usepackage[normalem]{ulem}
\usepackage{subcaption}
\usepackage{caption}
\usepackage{url}

\def\BibTeX{{\rm B\kern-.05em{\sc i\kern-.025em b}\kern-.08em
    T\kern-.1667em\lower.7ex\hbox{E}\kern-.125emX}}

\begin{document}


\title{
Visual Reasoning Evaluation of Grok, Deepseek’s Janus, Gemini, Qwen, Mistral, and ChatGPT
}










\author[1 2]{Nidhal Jegham}
\author[2, 3]{Marwan Abdelatti}
\author[2]{Abdeltawab Hendawi}

\affil[1]{Tunis Business School, University of Tunis, Tunis, Tunisia}
\affil[1]{\textit{jeghamnidhal7@gmail.com}}
\affil[2]{Computer Science Dept. University of Rhode Island, RI, USA}
\affil[2]{\textit{nidhal.jegham@uri.edu}}

\affil[2]{\textit{mabdelrazik@uri.edu}}
\affil[2]{\textit{hendawi@uri.edu}}

\affil[2]{Department of Computer Science Providence College, RI, USA}
\affil[3]{\textit{mabdelat@providence.edu}}

\maketitle

\begin{abstract}

Traditional evaluations of multimodal large language models (LLMs) have been limited by their focus on single-image reasoning, failing to assess crucial aspects like contextual understanding, reasoning stability, and uncertainty calibration.  This study addresses these limitations by introducing a novel benchmark that uniquely integrates multi-image reasoning tasks with rejection-based evaluation and positional bias detection. To rigorously evaluate these dimensions, we further introduce entropy as a novel metric for quantifying reasoning consistency across reordered answer variants. We applied this innovative benchmark to assess Grok 3, ChatGPT-4o, ChatGPT-o1, Gemini 2.0 Flash Experimental, DeepSeek’s Janus models, Qwen2.5-VL-72B-Instruct, QVQ-72B-Preview, and Pixtral 12B across eight visual reasoning tasks, including difference spotting and diagram interpretation. Our findings reveal ChatGPT-o1 leading in overall accuracy (82.5\%) and rejection accuracy (70.0\%), closely followed by Gemini 2.0 Flash Experimental (70.8\%). QVQ-72B-Preview demonstrated superior rejection accuracy (85.5\%). Notably, Pixtral 12B (51.7\%) showed promise in specific domains, while Janus models exhibited challenges in bias and uncertainty calibration, reflected in low rejection accuracies and high entropy scores.  High entropy scores in Janus models (Janus 7B: 0.8392, Janus 1B: 0.787) underscore their susceptibility to positional bias and unstable reasoning, contrasting with the low entropy and robust reasoning of ChatGPT models.  The study further demonstrates that model size is not the sole determinant of performance, as evidenced by Grok 3's underperformance despite its substantial parameter count. By employing multi-image contexts, rejection mechanisms, and entropy-based consistency metrics, this benchmark sets a new standard for evaluating multimodal LLMs, enabling a more robust and reliable assessment of next-generation AI systems.

\end{abstract}

\begin{IEEEkeywords}
Grok 3, Deepseek, Janus, Gemini 2.0 Flash, ChatGPT-4o, ChatGPT-o1, Mistral, Benchmark, Visual Reasoning, Evaluation, 
\end{IEEEkeywords}

\section{Introduction}

The rapid advancement of large language models (LLMs) has transformed artificial intelligence, enabling state-of-the-art performance in natural language processing (NLP), reasoning, and generative tasks \cite{brown2020language, chowdhery2022palm, touvron2023llama}. While early LLMs were primarily trained on text-based data, recent developments have led to the emergence of multimodal LLMs, capable of processing and reasoning over diverse data modalities, including text, images, videos, and structured information \cite{liu2023visual} \cite{chen2024mistral, openai2024gpt4o}. These models integrate vision and language understanding, significantly enhancing their applications across various domains, such as visual question answering (VQA) \cite{antol2015vqa, lu2022learn}, document comprehension \cite{mathew2021docvqa, huang2023layoutlm}, medical image interpretation \cite{wang2023biomedclip, mohta2023biogpt}, and multimodal dialogue systems \cite{shuster2021multi, zhu2024chat}.

Several multimodal models have demonstrated impressive capabilities in bridging vision and language understanding. $\mathbb{X}$AI's Grok 3 \cite{xai2025grok3} introduces advanced multimodal reasoning with a large parameter count, aiming to enhance contextual understanding and consistency. OpenAI’s GPT-4o \cite{openai2024gpt4o} extends its predecessor’s capabilities by incorporating image processing and reasoning over complex visual inputs. Google DeepMind’s Gemini 2.0 \cite{deepmind2024gemini} also advances multimodal interactions by integrating video and spatial reasoning. Meta’s LLaVA \cite{liu2023visual} aligns language and vision for improved visual grounding and generative capabilities, while Mistral's Pixtral 12B \cite{agrawal2024pixtral} introduces high-resolution image reasoning. In addition, open-source models such as DeepSeek-VL Janus \cite{deepseek2024vl} and Qwen-VL \cite{bai2023qwen} are pushing multimodal AI research forward, democratizing access to powerful vision-language models.

Evaluating multimodal LLMs remains a significant challenge, as existing benchmarks often assess isolated perception tasks rather than the complex reasoning required for real-world applications. While early datasets such as VQAv2 \cite{goyal2017making} and AI2D \cite{kembhavi2016diagram} focused on single-image understanding, recent benchmarks like NLVR2 \cite{suhr2019nlvr2}, MMMU \cite{yue2024mmmu}, and MathVista \cite{mathvista} have introduced multi-image tasks, logical comparisons, and mathematical reasoning. Moreover, MUIRBench \cite{wang2024muirbench} further advanced the field by integrating unanswerable question variants into multimodal visual reasoning. However, these efforts still fall short in systematically evaluating reasoning consistency, uncertainty calibration, and bias susceptibility. Addressing these gaps is crucial to ensuring that multimodal models move beyond pattern recognition and heuristic shortcuts to demonstrate genuine comprehension. 

 Unlike previous benchmarks, this study extends on the MUIRBench benchmark \cite{wang2024muirbench} and systematically evaluates overall reasoning capabilities, reasoning stability, rejection-based reasoning, and bias susceptibility by: (i) Reordering answer choices to assess whether models rely on heuristic-driven shortcuts rather than content understanding; (ii) introducing entropy as a novel metric to quantify reasoning stability and consistency across reordered variants, allowing for the detection of positional biases and randomness in answer selection; and (iii) testing rejection-based reasoning and abstention rate to measure whether models correctly abstain from answering when no valid option is provided.

The study evaluates multiple state-of-the-art multimodal LLMs, including Grok 3 \cite{xai2025grok3}, ChatGPT-o1 and ChatGPT-4o \cite{chatgpt}, Gemini 2.0 Flash Experimental \cite{gemini}, DeepSeek’s Janus models \cite{chen2025janus}, Pixtral 12B \cite{agrawal2024pixtral}, and Qwen-VL models \cite{bai2023qwen} to analyze how well-trending models generalize across these reasoning challenges. The results reveal notable discrepancies in performance. ChatGPT-4o and ChatGPT-o1 consistently achieve higher accuracy and reasoning stability, while Janus 7B and Janus 1B exhibit poor accuracy and high entropy scores, indicating significant reasoning variability and susceptibility to positional biases. This suggests that Janus models rely more on surface-level patterns rather than genuine comprehension, highlighting the importance of entropy as a consistency metric in identifying unstable reasoning.

The remainder of this paper is structured as follows. Section \ref{sec:related_work} reviews related work on multimodal benchmarks and evaluation methodologies. Section \ref{sec:benchmark_setup} describes the dataset used, the evaluated models, the experimental setup, and the evaluation framework, including the integration of new evaluation metrics based on reordered answers and entropy for reasoning consistency. Section \ref{sec:results} presents the results of the evaluation, highlighting key trends in accuracy, reasoning consistency, and rejection-based decision-making. Section \ref{sec:discussion} explores the broader implications of the findings, model-specific weaknesses, and improvements for multimodal AI evaluation. Finally, Section \ref{sec:conclusion} provides the conclusion and outlines future directions for refining multimodal benchmarking and reasoning assessment.

\section{Related Work} \label{sec:related_work}

\subsection{Visual Reasoning and Multi-Image Understanding Benchmarks}
Early benchmarks, such as VQAv2 \cite{goyal2017making}, GQA \cite{hudson2019gqa}, and AI2D \cite{kembhavi2016diagram}, laid the groundwork for evaluating multimodal models through single-image visual question answering (VQA). However, these datasets primarily assess perception and image-text alignment rather than deeper reasoning across multiple images.

To address this, recent benchmarks have expanded to multi-image contexts. NLVR2 \cite{suhr2019nlvr2} introduced logical comparisons between paired images, while MMMU \cite{yue2024mmmu} and MathVista \cite{mathvista} incorporated mathematical and diagrammatic reasoning using multiple visual inputs. MINT \cite{wang2023mint} and BLINK \cite{fu2024blink} further extended these evaluations by integrating temporal and spatial reasoning. Despite these improvements, these benchmarks do not explicitly evaluate reasoning consistency, answer stability, or a model’s ability to reject unanswerable questions, leaving a gap in assessing robustness in real-world scenarios.

Recent benchmarks such as SEED-Bench \cite{li2023seedbench} and ChartQA \cite{masry2022chartqa} have introduced confidence-based evaluation, but their focus remains limited to generative comprehension rather than multi-image, multi-choice answer stability. Meanwhile, MMLU-Math \cite{mathmmlu2023} and ReClor \cite{yu2020reclor} analyze answer order biases in textual reasoning, yet they do not extend this approach to multimodal contexts.

MUIRBench \cite{wang2024muirbench} represents a significant advancement in multi-image visual reasoning by incorporating an unanswerable question framework to evaluate rejection-based reasoning, testing whether models can recognize when no correct answer exists. However, MUIRBench does not incorporate reordered answer variations, making it difficult to determine whether models rely on positional heuristics rather than genuine comprehension. Without testing reasoning consistency across reordering, models may appear competent in fixed-choice settings while failing to generalize across structurally altered answer formats. Addressing these limitations is essential for evaluating multimodal models’ robustness and ensuring they do not exploit shortcuts in answer positioning rather than engaging in true visual reasoning.

\subsection{Reordered Answers, Positional Bias, and Rejection-Based Evaluation}
One of the overlooked weaknesses in current multimodal benchmarks is the reliance of models on heuristic patterns or randomness rather than true comprehension. This is particularly evident in the positional bias of multiple-choice answers, where models may favor specific answer choices based on their order rather than reasoning through the question. While text-based evaluations such as ReClor \cite{yu2020reclor} and MMLU-Math \cite{mathmmlu2023} incorporate reordered answers to test robustness, no major benchmark systematically applies this technique in multi-image reasoning.

To provide a more comprehensive assessment, our benchmark builds on MUIRBench by incorporating four key criteria: (i) Multi-image reasoning, (ii) unanswerable question recognition, (iii) reordered answer variations, and (iv) entropy-based reasoning consistency. In particular, we introduce entropy as a novel metric to measure consistency in responses across reordered versions of the same question. By quantifying variability in answer distributions, entropy identifies models that exhibit instability in reasoning, revealing reliance on positional heuristics or superficial patterns rather than genuine content understanding. Moreover, our benchmark evaluates rejection accuracy, determining whether models correctly abstain from answering when no valid option exists.

By integrating multi-image reasoning, reordered answer variations, entropy-based reasoning consistency, and rejection-based evaluation, this benchmark contributes a novel framework for diagnosing model weaknesses and guiding future advancements in multimodal LLMs and visual reasoning benchmarks. This dual approach of reordered answers and entropy sets a new standard for reasoning consistency evaluation, ensuring more robust, reliable, and generalizable AI systems.

\section{Benchmark Setup} \label{sec:benchmark_setup}

\subsection{Dataset}
To evaluate the multi-image reasoning capabilities of multimodal LLMs, we curated a subset of 120 questions and 376 images from the MUIRBench \cite{wang2024muirbench} dataset, ensuring a balanced assessment across diverse reasoning tasks. Unlike traditional benchmarks that focus on single-image tasks, this dataset challenges models to process and integrate multiple visual inputs, making it a more rigorous evaluation framework for contextual, spatial, and logical reasoning.

\begin{figure*}
    \centering
    \includegraphics[width=16cm]{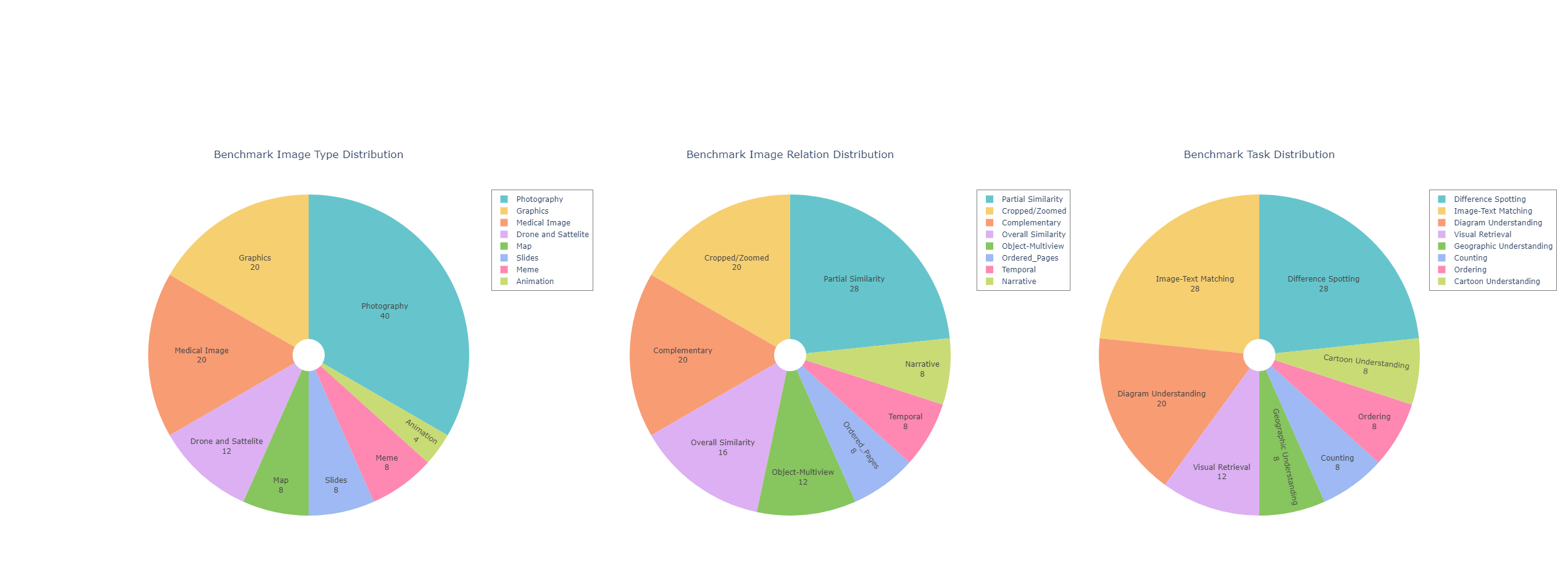}
    \caption{The distribution of the questions based on the type of images, image relations, and tasks.}
    \vspace{-1em}
    \label{fig:image_prop}
\end{figure*}

The dataset maintains diversity in image types, as illustrated in Figure \ref{fig:image_prop}, spanning real-world photographs, medical imagery, scientific diagrams, and satellite views. By integrating both curated and synthetic data, the benchmark mitigates data contamination risks, ensuring models are tested on truly novel inputs.

Additionally, the dataset includes a range of multi-image relations, depicted in Figure \ref{fig:image_prop}, challenging models to process various dependencies such as temporal sequences, complementary perspectives, and multi-view representations. This structured approach ensures that models are tested beyond simple pattern recognition, requiring advanced spatial, logical, and comparative reasoning.

The selection of questions from MUIRBench was guided by the need for a balanced and diverse evaluation while maintaining computational feasibility. Given that MUIRBench consists of 2,600 multiple-choice questions across 12 multi-image reasoning tasks, we curated a representative subset of 120 questions and 376 images, averaging 3.13 images per question. This selection ensures an efficient assessment of multimodal LLMs while focusing on key reasoning abilities.

We prioritized coverage of core multi-image tasks, selecting 8 diverse tasks that span both low-level perception (e.g., difference spotting) and high-level reasoning (e.g., image-text matching) as seen in Fig \ref{fig:image_prop}. Tasks with high significance in multimodal evaluation, such as Image-Text Matching and Difference Spotting, were given higher representation, each comprising 28 questions. Additionally, we ensured diversity in multi-image relationships, including temporal sequences, complementary perspectives, and multi-view representations, to evaluate models' ability to integrate spatial, logical, and comparative reasoning across multiple images.

Our selection also maintained MUIRBench’s answerable-unanswerable pairing strategy by including 40 unanswerable questions. This prevents models from exploiting shortcut biases and lucky guesses and encourages a deeper understanding of visual content. 

Additionally, we incorporated alternate versions of each question by reordering the answer choices. This ensures that models do not rely on positional biases or heuristic patterns but genuinely understand and reason the multi-image context. By shuffling the answer choices, we can verify whether a model is consistently selecting the correct response rather than guessing based on positional tendencies. 

This question refinement strategy improves model robustness by ensuring they rely on reasoning rather than heuristics. Our study offers a rigorous and efficient way to test how well next-generation vision-language models can reason about multiple images.

\subsection{Models}

In our evaluation, we utilized a diverse set of multimodal language models, as seen in Table \ref{tab: llm_table}, each designed with unique capabilities to process and reason over visual inputs. Below is a brief overview of each model: 

\subsubsection{Janus 7B} 
Developed by DeepSeek and was released on January 27, 2025, Janus 7B \cite{chen2025janus} is a 7-billion parameter open-source multimodal model built for advanced image understanding and text-to-image generation. It supports multiple-image processing, making it suitable for complex visual reasoning tasks.

\subsubsection{Janus 1B} 
developed by DeepSeek and was Released alongside Janus 7B on January 27, 2025. Janus 1B \cite{chen2025janus} is a lightweight 1-billion parameter open-source variant, designed for efficient multi-image processing while maintaining a lower computational footprint.

\subsubsection{Grok 3} Grok 3 is a multimodal large language model developed by $\mathbb{X}$AI. Released on February 17, 2025, Grok 3 boasts 2.7 trillion parameters, making it one of the most advanced AI models to date. 

\subsubsection{Gemini 2.0 Flash Experimental} 
Launched on February 10, 2025, Gemini 2.0 Flash Experimental \cite{gemini} is an advanced multimodal model developed by DeepMind that processes both multiple images and videos. It is optimized for rapid inference and efficient memory usage, making it well-suited for real-time visual reasoning applications.

\subsubsection{QVQ-72B-Preview} 
Released on December 25, 2024, QVQ-72B-Preview \cite{bai2023qwen} is a 72-billion parameter open-source vision-language model developed by Alibaba that introduces novel techniques in visual question answering. It supports multi-image reasoning, allowing for better contextual understanding across images.

\subsubsection{Qwen2.5-VL-72B-Instruct} 
Developed by Alibaba and released on September 19, 2024, Qwen2.5-VL-72B-Instruct \cite{bai2023qwen} is a 72-billion parameter open-source instruction-tuned vision-language model. It incorporates dynamic resolution handling and multimodal rotary position embedding (M-ROPE), enabling multi-image and video comprehension.

\subsubsection{Pixtral 12B} 
Released on September 17, 2024, Pixtral 12B \cite{agrawal2024pixtral} is a 12-billion parameter open-source multimodal model developed by Mistral AI, specializing in high-resolution image analysis and text generation.

\subsubsection{ChatGPT-o1} 
Introduced on December 5, 2024, ChatGPT-o1 \cite{chatgpt} is an advanced iteration of OpenAI’s multimodal models, supporting multiple images and video inputs. It enhances contextual multimodal understanding, making it effective for vision-language interactions.

\subsubsection{ChatGPT-4o} 
Released on May 13, 2024, ChatGPT-4o \cite{chatgpt} is a multimodal model developed by OpenAI, capable of processing multiple images. It improves on its predecessors by enhancing image-text alignment and expanding multimodal reasoning capabilities.

\begin{table}[]
\caption{Summary of evaluated multimodal language models, including their parameter sizes, release dates, image processing capabilities, and open-source availability.}

\begin{tabular}{|l|c|c|c|c|}
\hline
\multicolumn{1}{|c|}{\textbf{Model}}                                        & \textbf{\begin{tabular}[c]{@{}c@{}}Number of\\ Parameters\end{tabular}} & \textbf{Release Date} & \textbf{\begin{tabular}[c]{@{}c@{}}Image \\ Processing \\ Capability\end{tabular}} & \textbf{\begin{tabular}[c]{@{}c@{}}Open\\ Source\end{tabular}} \\ \hline
Janus 7B                                                                    & 7B                                                                      & Jan 27, 2025          & \begin{tabular}[c]{@{}c@{}}Multiple \\ images\end{tabular}                         & Yes                                                            \\ \hline
Janus 1B                                                                    & 1B                                                                      & Jan 27, 2025          & \begin{tabular}[c]{@{}c@{}}Multiple\\ images\end{tabular}                          & Yes                                                            \\ \hline
Grok 3                                                                      & 2.7T                                                                    & Feb  17, 2025         & \begin{tabular}[c]{@{}c@{}}Multiple\\ images\end{tabular}                          & No                                                             \\ \hline
\begin{tabular}[c]{@{}l@{}}Gemini 2.0 \\ Flash \\ Experimental\end{tabular} & N/A                                                                     & Feb 10, 2025          & \begin{tabular}[c]{@{}c@{}}Multiple \\ images\\ and videos\end{tabular}            & No                                                             \\ \hline
\begin{tabular}[c]{@{}l@{}}QVQ-72B-\\ Preview\end{tabular}                  & 72B                                                                     & Dec 25, 2024          & \begin{tabular}[c]{@{}c@{}}Multiple \\ images\end{tabular}                         & Yes                                                            \\ \hline
\begin{tabular}[c]{@{}l@{}}Qwen2.5-VL-\\ 72B-Instruct\end{tabular}          & 72B                                                                     & Sep 19, 2024          & \begin{tabular}[c]{@{}c@{}}Multiple\\ images\\ and videos\end{tabular}             & Yes                                                            \\ \hline
Pixtral 12B                                                                 & 12B                                                                     & Sep 17, 2024          & \begin{tabular}[c]{@{}c@{}}Multiple \\ images\end{tabular}                         & Yes                                                            \\ \hline
ChatGPT-o1                                                                  & N/A                                                                     & Dec 5, 2024           & \begin{tabular}[c]{@{}c@{}}Multiple\\ images\\ and videos\end{tabular}             & No                                                             \\ \hline
ChatGPT-4o                                                                  & N/A                                                                     & May 13, 2024          & \begin{tabular}[c]{@{}c@{}}Multiple \\ images\end{tabular}                         & No                                                             \\ \hline
\end{tabular}
\label{tab: llm_table}
\end{table}

This selection encompasses a range of models with varying capacities and innovations, providing a comprehensive basis for evaluating multimodal language models in visual reasoning tasks.

\section{Methodology}

\subsection{Experimental Setup}

\subsubsection{Question Presentation Format} 
To ensure a standardized evaluation across all models, each question follows a structured format where the model receives a textual prompt and the corresponding images. The format is as follows:

\begin{quote}
\textbf{Input:} \{Question\} These are the options: [A) First Option, B) Second Option, C) Third Option, D) Fourth.]
\end{quote}
\begin{quote}
\textbf{Example:} Which picture below better fits the description: A black and white cat sitting in a green bowl? These are the options: [A) First image, B) Second image, C) None of the choices provided, D) Third image.]
\end{quote}

Models are expected to select the most appropriate answer based on the given images. To assess model consistency and robustness, each question is tested in different variations:
\begin{enumerate}
    \item Answerable Version: The correct answer is available among the choices.
    \item Reordered Version: The same question with shuffled answer choices to detect positional biases.
    \item Unanswerable Version: The correct answer is removed or altered, testing the model’s ability to recognize when a question cannot be answered.
\end{enumerate}

\begin{figure}[h]
    \centering
    \includegraphics[width=0.9\linewidth]{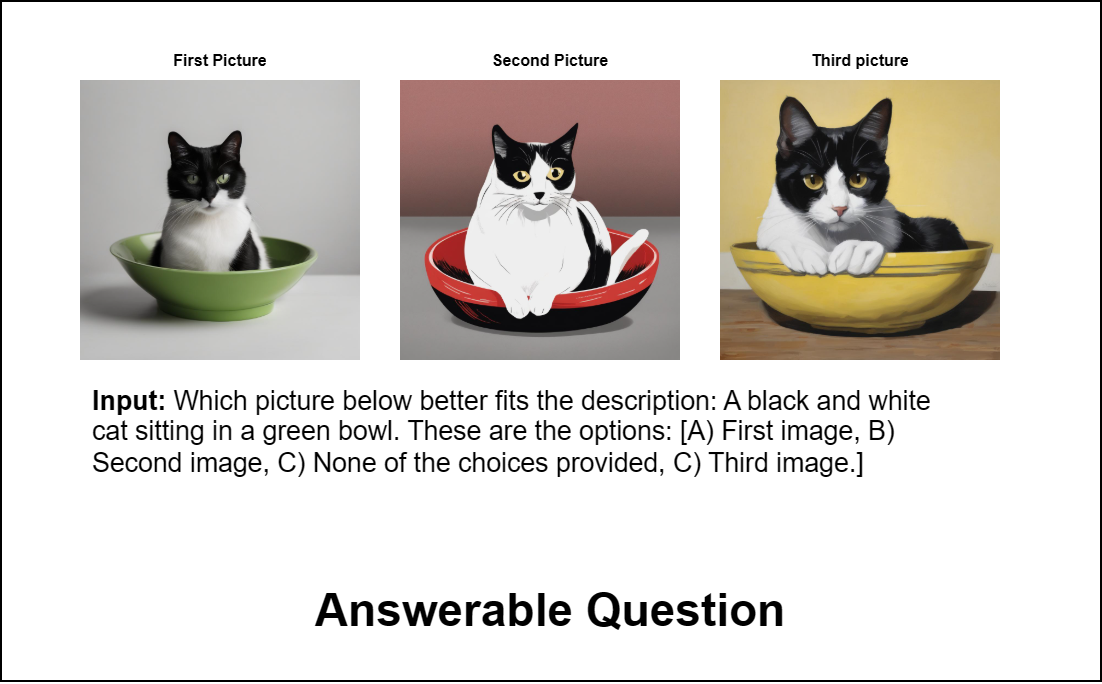}
    \caption{Example of an answerable question where the correct answer is available among the choices.}
    \label{fig:answerable}
\end{figure}

\begin{figure}[h]
    \centering
    \includegraphics[width=0.9\linewidth]{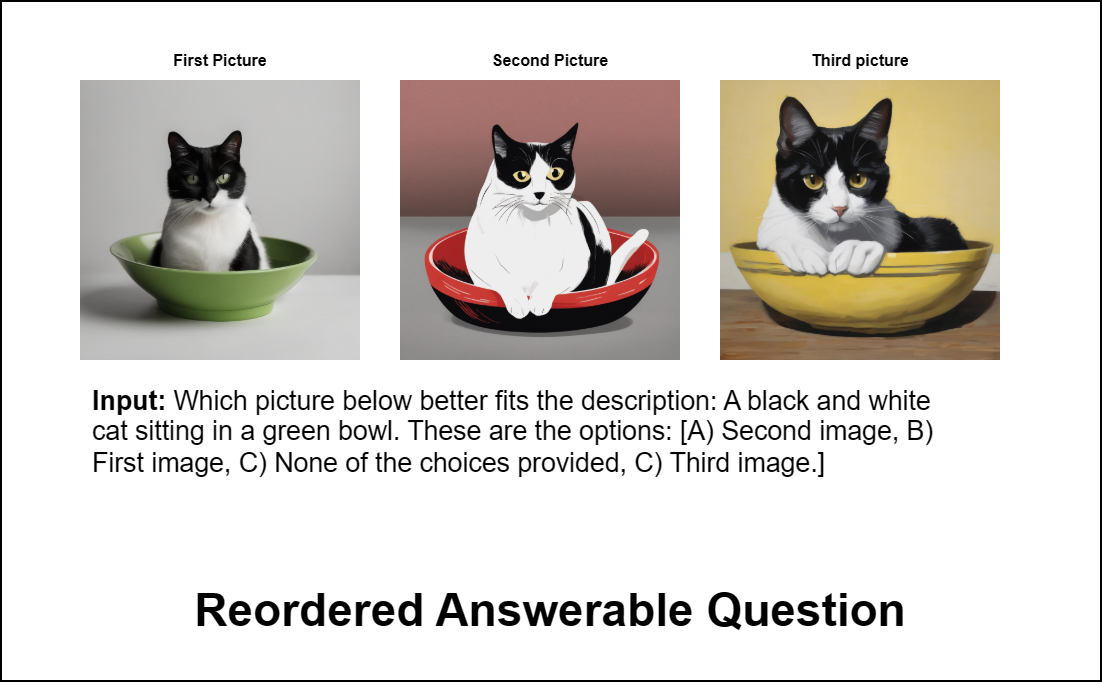}
    \caption{Example of the same question with reordered answer choices to test for positional biases.}
    \label{fig:reordered}
\end{figure}

\begin{figure}[h]
    \centering
    \includegraphics[width=0.9\linewidth]{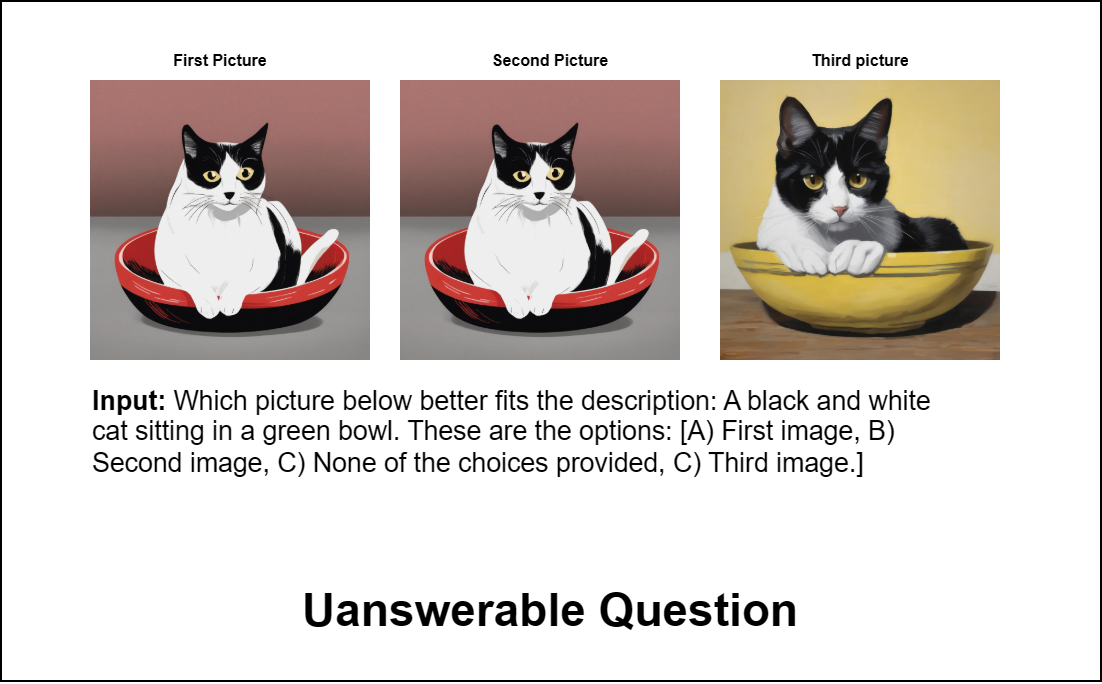}
    \caption{Example of an unanswerable question where the correct choice is removed or altered.}
    \label{fig:unanswerable}
\end{figure}

To further clarify this evaluation process, Figures \ref{fig:answerable}, \ref{fig:reordered}, and \ref{fig:unanswerable} illustrate examples of each variation.

\subsubsection{Local Deployment for Janus Models} 
For Janus 7B and Janus 1B, we ran the experiments on a high-performance workstation equipped with two NVIDIA RTX 4090 GPUs, each featuring 16,384 CUDA cores due to the unavailability of an official online interface.

\subsubsection{Web-Based Evaluation for Other Models} 
For the remaining models (Grok3, QVQ-72B-Preview, Qwen2.5-VL-72B-Instruct, Pixtral 12B, ChatGPT-4o, ChatGPT-o1, and Gemini 2.0 Flash Experimental), we conducted evaluations using their official online interfaces. To ensure zero-shot learning, a new chat session was initialized for each question, preventing any contextual memory retention. We set the temperature to 1.0, encouraging diverse yet unbiased model responses.

\subsubsection{Consistency and Fairness Measures}
All models were assessed using identical input formats, with questions presented in their original multiple-choice structure. No additional context was given beyond the provided question and images, ensuring models relied solely on intrinsic reasoning capabilities.

This setup guarantees a standardized comparison across models, enabling a robust assessment of multi-image reasoning capabilities.

\subsection{Evaluation Setup}

To ensure a standardized and fair assessment of all multimodal LLMs, we established a strict evaluation protocol that handles varying model constraints, answer formats, and consistency measures.

\subsubsection{Handling of Image Restrictions} 
Some models have restrictive policies that prevent them from processing images, either due to safety mechanisms or inherent limitations. If a model fails to accept an image input, the response is automatically marked as incorrect, as it indicates an inability to engage with the core visual reasoning task.

\subsubsection{Answer Parsing and Validity} 
To eliminate ambiguity in model responses, we adhere to a strict answer validation process:
\begin{itemize}
    \item If a model provides an answer choice that does not exist among the given options, the response is considered incorrect.
    \item If the model correctly identifies the answer but provides an incorrect explanation, the response is still marked as correct, as the primary evaluation metric is answer selection. For instance, if a model produces an answer such as “C) First image” when the correct answer is “C) Second image”, the response is marked as correct since the answer letter aligns with the expected choice.
\end{itemize}

\subsubsection{Ensuring Answer Consistency} 
To detect inconsistencies in model reasoning, we incorporate reordered answer choices across repeated questions. If a model guesses correctly on one instance but incorrectly on another reordered variant, this indicates reliance on positional heuristics rather than genuine understanding. This methodology allows us to identify unreliable answering patterns and assess whether a model is robust to answer order changes.

This evaluation setup ensures that models are judged on their actual reasoning abilities, eliminating potential issues or biases caused by format variations or restrictive policies.

\subsubsection{Overall Accuracy Evaluation}

Each correctly answered question is assigned a score of one point, with all questions weighed equally regardless of their difficulty or complexity. The evaluation metric is based on accuracy, defined as:

\begin{equation}
\text{Accuracy} = \frac{\text{Number of Correct Answers}}{\text{Total Number of Questions}} \times 100\%
\end{equation}

Accuracy is a fundamental metric in evaluating multimodal large language models (MLLMs), as it provides a straightforward measure of a model's performance across various tasks. For instance, the MME benchmark employs accuracy to assess both perception and cognition abilities of MLLMs across multiple subtasks \cite{MME2023}. Similarly, the MM-BigBench framework utilizes accuracy to evaluate model performance on a wide range of multimodal content comprehension tasks \cite{MMBigBench2023}.

\subsubsection{Rejection Accuracy Evaluation}

In addition to overall accuracy, we evaluate each model’s ability to recognize when no correct answer is available. Rejection accuracy measures how well a model correctly selects the ``None of the provided options" choice when no valid answer is present. This is a critical aspect of uncertainty calibration, ensuring that models do not make incorrect guesses when faced with unanswerable questions.

Rejection accuracy is computed as:

\begin{equation}
\text{Rejection Accuracy} = \frac{\text{Number of Correct Rejections}}{\text{Total Rejection Questions}} \times 100\%
\end{equation}

where ``Number of Correct Rejections" represents the cases where the model correctly abstains from selecting an incorrect answer, and ``Total Rejection Questions" corresponds to a predefined subset of 40 unanswerable questions in the benchmark.

Evaluating a model's ability to handle unanswerable questions is essential for deploying reliable AI systems. While traditional accuracy metrics assess performance on answerable queries, incorporating rejection accuracy provides a more comprehensive evaluation framework, as it accounts for a model's uncertainty calibration and decision-making in ambiguous scenarios. Previous studies, such as MUIRBench \cite{wang2024muirbench}, the comprehensive evaluation framework in ChEF \cite{ChEF2023}, and the survey on multimodal LLM evaluation \cite{EvaluationSurvey2023}, highlight the importance of rejection accuracy in assessing model robustness and reliability.

\subsubsection{Abstention Rate Evaluation}

In addition to overall accuracy and rejection accuracy, we evaluate each model's tendency to select ``None of the provided options" across all questions. This metric, termed Abstention Rate, measures the proportion of times a model opts for ``None of the provided options", reflecting its inclination to abstain from answering.

\begin{equation}
\text{Abstention Rate} = 
\frac{
    \begin{array}{c}
    \text{Number of ``None of the} \\
    \text{provided options" Selections}
    \end{array}
}{
    \text{Total Questions}
} \times 100\%
\end{equation}

A higher abstention rate indicates a model's cautious approach, potentially avoiding incorrect answers when uncertain, while a lower rate suggests overconfidence or reluctance to admit uncertainty.

This metric complements traditional accuracy and rejection accuracy by offering insights into a model's uncertainty calibration. Previous studies have explored similar concepts, emphasizing the importance of abstention in enhancing reliability and safety in LLMs \cite{wen2024abstention, madhusudhan2024abstention}.

\subsubsection{Entropy-Based Reasoning Consistency Evaluation} 

To quantify reasoning stability, we utilize Entropy as a metric to evaluate a model's consistency in answering reordered variations of the same question. Entropy captures the uncertainty and variability in the model’s responses across these variants, providing a robust measure of reasoning consistency.

Entropy is calculated by grouping each original question with its reordered variants and analyzing the distribution of the model's responses. Specifically, the frequency of each answer option is computed and normalized to form a probability distribution while the correct answer is unchanged throughout all the question variants. Entropy is then calculated using the formula:

\begin{equation}
    H(Q_i) = - \sum_{j=1}^{k} p(a_j) \log_2(p(a_j))
    \label{eq:entropy}
\end{equation}

where \( H(Q_i) \) is the entropy for question group \( i \), \( k \) is the total number of possible answer options, and \( p(a_j) \) is the probability of selecting option \( a_j \) among the reordered variants.

The Mean Entropy for a model across all questions is computed as:

\begin{equation}
    \text{Mean Entropy}_m = \frac{1}{N} \sum_{i=1}^{N} H(Q_i)
    \label{eq:mean_entropy}
\end{equation}

where \( \text{Mean Entropy}_m \) represents the average reasoning consistency for model \( m \), and \( N \) is the total number of question groups. 

Lower entropy values indicate higher consistency, suggesting that the model consistently selects the same answer across reordered variants, reflecting stable reasoning. Conversely, higher entropy values suggest greater variability in answers, indicating potential instability or influence from positional biases rather than consistent reasoning patterns.

Previous studies have utilized entropy-based metrics to evaluate reasoning consistency and adaptive choice behavior in LLMs, such as in information selection for Chain-of-Thought prompting \cite{inform_entropy_reasoning} and in data compression proficiency assessments \cite{matrix_entropy_llm}. These approaches mainly focus on text generation tasks, where entropy is used to adjust dynamic temperature sampling or to assess information selection efficiency. However, these methods do not address reasoning consistency across reordered variants in multimodal visual reasoning. 

Our approach applies entropy to evaluate consistency specifically across reordered variants, which allows us to measure reasoning stability and detect positional biases that may influence model outputs. This offers a novel perspective on robustness and stability in multimodal LLMs, particularly in complex visual contexts where reasoning consistency is crucial for accurate interpretation. By assessing entropy across answer distributions for reordered variants, our method provides a more comprehensive understanding of reasoning patterns and positional biases compared to traditional accuracy metrics.

\section{results}\label{sec:results}

\begin{figure*}[h]
    \centering
    \includegraphics[width=\linewidth]{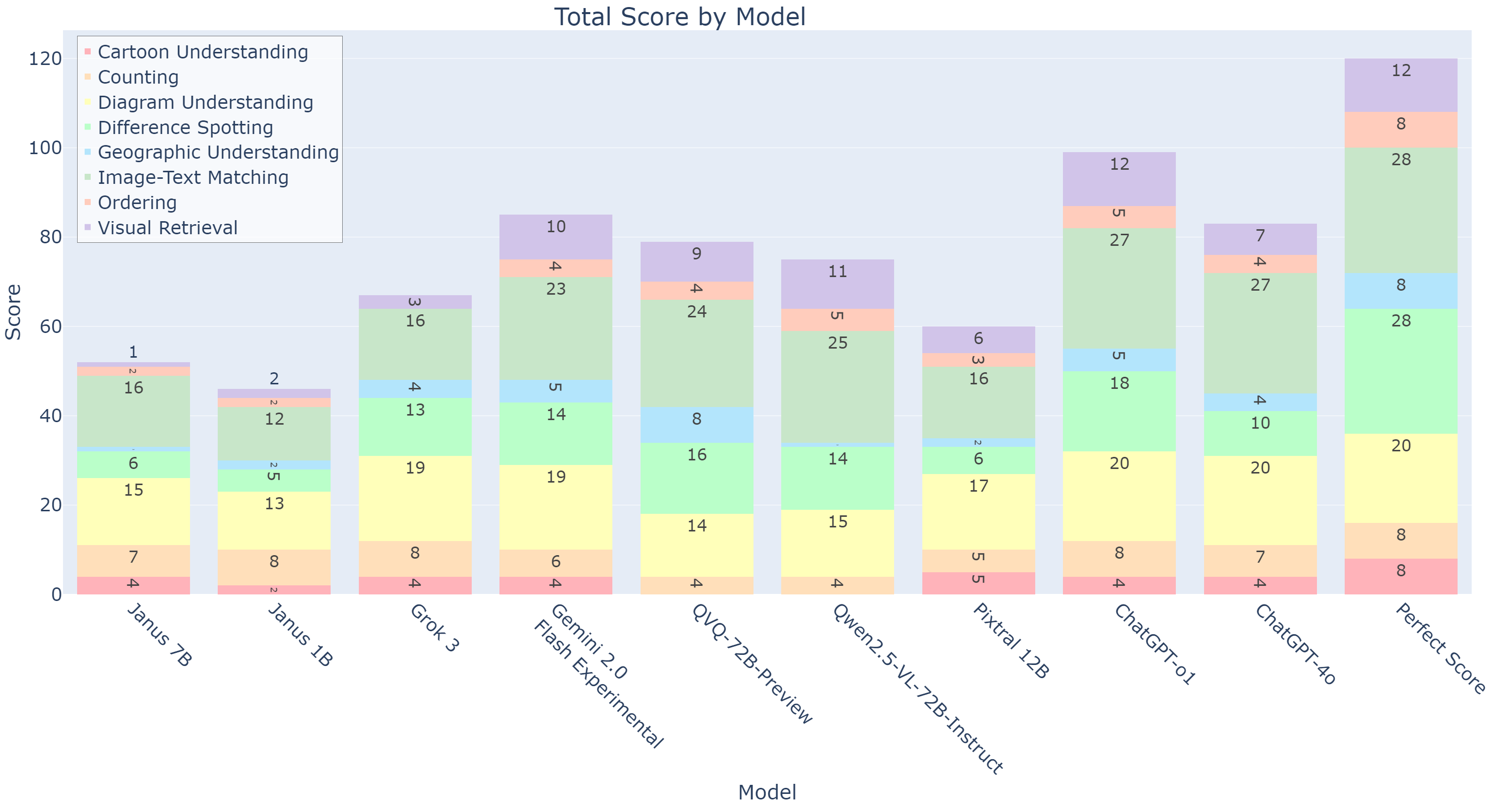}
    \caption{total accuracy results of each model. The ``Perfect Score'' represents the ideal scenario where the model answers all the questions correctly.}
    \label{fig:total_score}
\end{figure*}

\begin{figure}[h]
    \centering
    \includegraphics[width=0.8\linewidth]{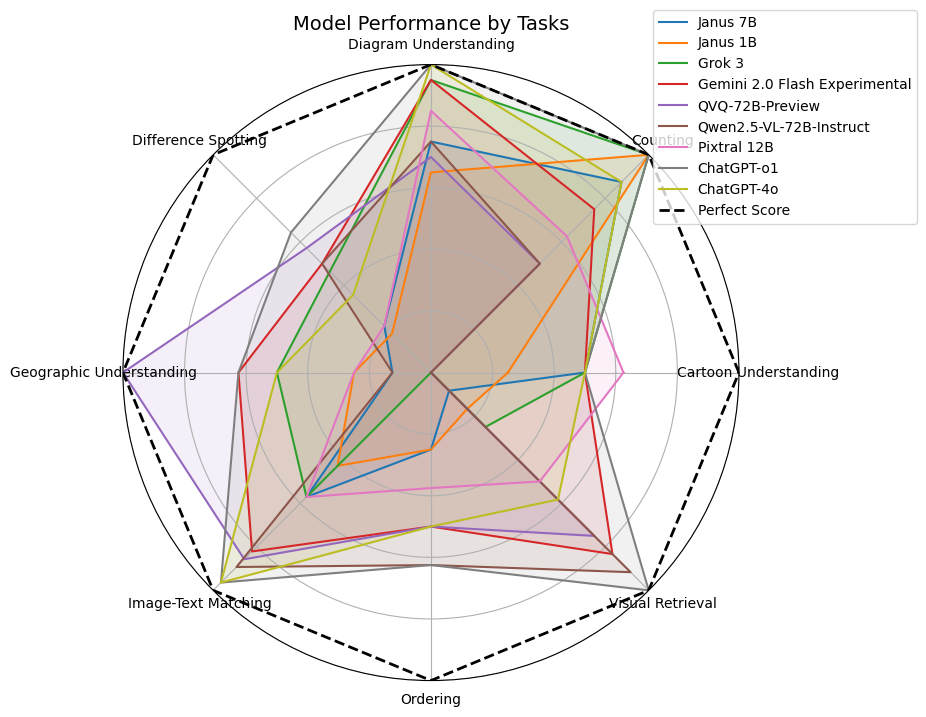}
    \caption{Radar chart visualizing model performance across all tasks.}
    \label{fig:radar_chart}
\end{figure}

\begin{table*}[]
\caption{Accuracy of each model across tasks}

\begin{tabular}{|l|l|l|l|l|l|l|l|l|l|}
\hline
\textbf{Model}                                                          & \textbf{\begin{tabular}[c]{@{}l@{}}Cartoon\\ Understanding\end{tabular}} & \textbf{Counting} & \textbf{\begin{tabular}[c]{@{}l@{}}Diagram \\ Understanding\end{tabular}} & \textbf{\begin{tabular}[c]{@{}l@{}}Difference \\ Spotting\end{tabular}} & \textbf{\begin{tabular}[c]{@{}l@{}}Geographic \\ Understanding\end{tabular}} & \textbf{\begin{tabular}[c]{@{}l@{}}Image-Text \\ Matching\end{tabular}} & \textbf{Ordering} & \textbf{\begin{tabular}[c]{@{}l@{}}Visual \\ Retrieval\end{tabular}} & \textbf{Total} \\ \hline
Janus 7B                                                                & 0.5                                                                      & 0.875             & 0.75                                                                      & 0.214                                                                   & 0.125                                                                        & 0.5714                                                                  & 0.25              & 0.0833                                                               & 0.433          \\ \hline
Janus 1B                                                                & 0.25                                                                     & \textbf{1.0}      & 0.65                                                                      & 0.1785                                                                  & 0.25                                                                         & 0.4285                                                                  & 0.25              & 0.166                                                                & 0.3833         \\ \hline
Grok 3                                                                  & 0.5                                                                      & \textbf{1.0}      & 0.95                                                                      & 0.4642                                                                  & 0.5                                                                          & 0.5714                                                                  & 0.0               & 0.25                                                                 & 0.558          \\ \hline
\begin{tabular}[c]{@{}l@{}}Gemini 2.0 Flash\\ Experimental\end{tabular} & 0.5                                                                      & 0.75              & 0.95                                                                      & 0.5                                                                     & \textbf{0.625}                                                               & 0.8214                                                                  & 0.5               & 0.833                                                                & 0.7083         \\ \hline
\begin{tabular}[c]{@{}l@{}}QVQ-72B\\ -Preview\end{tabular}              & 0.0                                                                      & 0.5               & 0.7                                                                       & 0.5714                                                                  & 1.0                                                                          & 0.8571                                                                  & 0.5               & 0.75                                                                 & 0.6583         \\ \hline
\begin{tabular}[c]{@{}l@{}}Qwen2.5-VL\\ -72B-Instruct\end{tabular}      & 0.0                                                                      & 0.5               & 0.75                                                                      & 0.5                                                                     & 0.125                                                                        & 0.8928                                                                  & \textbf{0.625}    & 0.916                                                                & 0.625          \\ \hline
Pixtral 12B                                                             & \textbf{0.625}                                                           & 0.625             & 0.85                                                                      & 0.2142                                                                  & 0.25                                                                         & 0.5714                                                                  & 0.375             & 0.5                                                                  & 0.5            \\ \hline
ChatGPT-o1                                                              & 0.5                                                                      & \textbf{1.0}      & \textbf{1.0}                                                              & \textbf{0.6428}                                                         & \textbf{0.625}                                                               & \textbf{0.9642}                                                         & \textbf{0.625}    & \textbf{1.0}                                                         & \textbf{0.825} \\ \hline
ChatGPT-4o                                                              & 0.5                                                                      & 0.875             & \textbf{1.0}                                                              & 0.3571                                                                  & 0.5                                                                          & \textbf{0.9642}                                                         & 0.5               & 0.5833                                                               & 0.6916         \\ \hline
\end{tabular}

\label{tab:accuracy_score}

\end{table*}

\begin{table*}[]
\caption{Number of correctly answered questions per model across tasks.}
\begin{tabular}{|l|l|l|l|l|l|l|l|l|l|}
\hline
\textbf{Model}                                                           & \textbf{\begin{tabular}[c]{@{}l@{}}Cartoon \\ Understanding\end{tabular}} & \textbf{Counting} & \textbf{\begin{tabular}[c]{@{}l@{}}Diagram\\ Understanding\end{tabular}} & \textbf{\begin{tabular}[c]{@{}l@{}}Difference\\  Spotting\end{tabular}} & \textbf{\begin{tabular}[c]{@{}l@{}}Geographic\\  Understanding\end{tabular}} & \textbf{\begin{tabular}[c]{@{}l@{}}Image-Text\\  Matching\end{tabular}} & \textbf{Ordering} & \textbf{\begin{tabular}[c]{@{}l@{}}Visual\\  Retrieval\end{tabular}} & \textbf{Total} \\ \hline
Janus 7B                                                                 & 4                                                                         & 7                 & 15                                                                       & 6                                                                       & 1                                                                            & 16                                                                      & 2                 & 1                                                                    & 52             \\ \hline
Janus 1B                                                                 & 2                                                                         & 8                 & 13                                                                       & 5                                                                       & 2                                                                            & 12                                                                      & 2                 & 2                                                                    & 46             \\ \hline
Grok 3                                                                   & 4                                                                         & 8                 & 19                                                                       & 13                                                                      & 4                                                                            & 16                                                                      & 0                 & 3                                                                    & 67             \\ \hline
\begin{tabular}[c]{@{}l@{}}Gemini 2.0 Flash\\  Experimental\end{tabular} & 4                                                                         & 6                 & 19                                                                       & 14                                                                      & 5                                                                            & 23                                                                      & 4                 & 10                                                                   & 85             \\ \hline
\begin{tabular}[c]{@{}l@{}}QVQ-72B-\\ Preview\end{tabular}               & 0                                                                         & 4                 & 14                                                                       & 16                                                                      & \textbf{8}                                                                   & 24                                                                      & 4                 & 9                                                                    & 79             \\ \hline
\begin{tabular}[c]{@{}l@{}}Qwen2.5-VL\\ -72B-Instruct\end{tabular}       & 0                                                                         & 4                 & 15                                                                       & 14                                                                      & 1                                                                            & 25                                                                      & \textbf{5}        & 11                                                                   & 75             \\ \hline
Pixtral 12B                                                              & \textbf{5}                                                                & 5                 & 17                                                                       & 6                                                                       & 2                                                                            & 16                                                                      & 3                 & 6                                                                    & 60             \\ \hline
ChatGPT-o1                                                               & 4                                                                         & \textbf{8}        & \textbf{20}                                                              & \textbf{18}                                                             & 5                                                                            & \textbf{27}                                                             & \textbf{5}        & \textbf{12}                                                          & \textbf{99}    \\ \hline
ChatGPT-4o                                                               & 4                                                                         & 7                 & \textbf{20}                                                              & 10                                                                      & 4                                                                            & \textbf{27}                                                             & 4                 & 7                                                                    & 83             \\ \hline
Number of Questions                                                      & 8                                                                         & 8                 & 20                                                                       & 28                                                                      & 8                                                                            & 28                                                                      & 8                 & 12                                                                   & 120            \\ \hline
\end{tabular}
\label{tab:answer_score}
\end{table*}

\begin{table}[h]
\centering
\caption{Rejection Accuracy and Abstention Rate of each model across tasks}
\label{tab:none_answer_accuracy}
\begin{tabular}{|l|l|l|l|l|}
\hline
\textbf{Model}                                                          & \textbf{\begin{tabular}[c]{@{}l@{}}Correct \\ Answers\end{tabular}} & \textbf{\begin{tabular}[c]{@{}l@{}}Accuracy \\ (\%)\end{tabular}} & \textbf{\begin{tabular}[c]{@{}l@{}}Abstention \\ Answer \\ Count\end{tabular}} & \textbf{\begin{tabular}[c]{@{}l@{}}Abstention\\  Rate\end{tabular}} \\ \hline
Janus 7B                                                                & 6                                                                   & 0.15                                                              & 12                                                                             & 0.1                                                                 \\ \hline
Janus 1B                                                                & \textbf{2}                                                          & \textbf{0.05}                                                     & 5                                                                              & 0.041                                                               \\ \hline
Grok 3                                                                  & 21                                                                  & 0.525                                                             & 45                                                                             & 0.375                                                               \\ \hline
\begin{tabular}[c]{@{}l@{}}Gemini 2.0 Flash\\ Experimental\end{tabular} & 20                                                                  & 0.5                                                               & 26                                                                             & 0.216                                                               \\ \hline
\begin{tabular}[c]{@{}l@{}}QVQ-72B-\\ Preview\end{tabular}              & \textbf{34}                                                         & \textbf{0.855}                                                    & \textbf{51}                                                                    & \textbf{0.425}                                                      \\ \hline
\begin{tabular}[c]{@{}l@{}}Qwen2.5-VL-\\ 72B-Instruct\end{tabular}      & 21                                                                  & 0.525                                                             & 33                                                                             & 0.275                                                               \\ \hline
Pixtral 12B                                                             & 12                                                                  & 0.3                                                               & 18                                                                             & 0.15                                                                \\ \hline
ChatGPT-o1                                                              & 28                                                                  & 0.7                                                               & 32                                                                             & 0.267                                                               \\ \hline
ChatGPT-4o                                                              & 18                                                                  & 0.450                                                             & 25                                                                             & 0.208                                                               \\ \hline
Perfect Score                                                           & 40                                                                  & 1                                                                 & 40                                                                             & 0.334                                                               \\ \hline
\end{tabular}
\end{table}

\begin{table}[h]
\centering
\caption{Entropy-Based Reasoning Consistency Scores}
\label{tab:miqv}
\begin{tabular}{|l|l|}
\hline
\textbf{Model}                                                          & \textbf{Entropy} \\ \hline
Janus 7B                                                                & \textbf{0.8392}  \\ \hline
Janus 1B                                                                & 0.787            \\ \hline
Grok 3                                                                  & 0.256            \\ \hline
\begin{tabular}[c]{@{}l@{}}Gemini 2.0 Flash\\ Experimental\end{tabular} & 0.3163           \\ \hline
\begin{tabular}[c]{@{}l@{}}QVQ-72B-\\ Preview\end{tabular}              & 0.3537           \\ \hline
\begin{tabular}[c]{@{}l@{}}Qwen2.5-VL-\\ 72B-Instruct\end{tabular}      & 0.4892           \\ \hline
Pixtral 12B                                                             & 0.557            \\ \hline
ChatGPT-o1                                                              & \textbf{0.1352}  \\ \hline
ChatGPT-4o                                                              & 0.216            \\ \hline
\end{tabular}
\end{table}

\begin{figure*}[h]
    \centering
    \includegraphics[width=\linewidth]{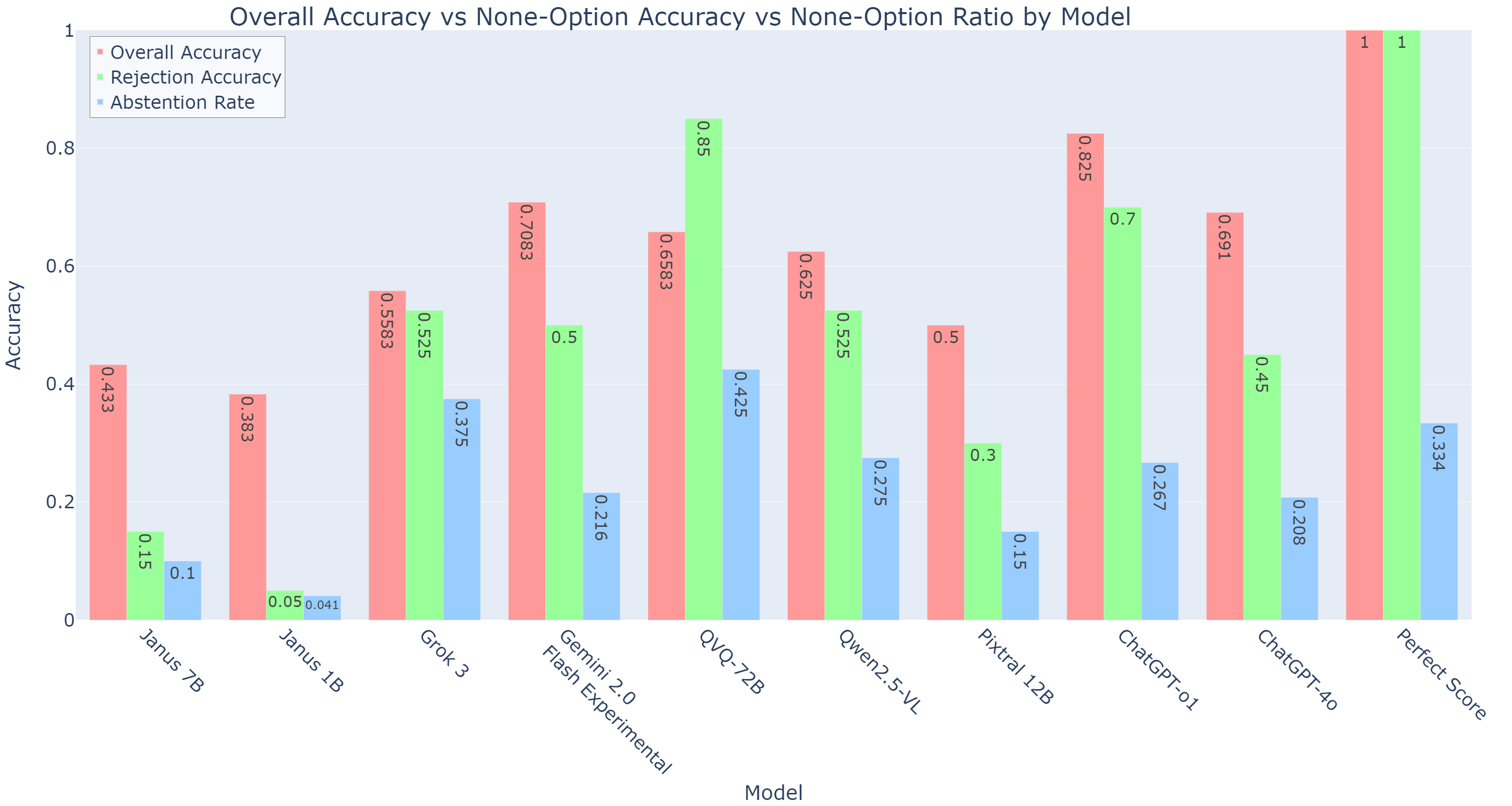}
    \caption{Comparison of total accuracy and rejection accuracy for each model. The ``Perfect Score'' represents the ideal scenario with 100\% overall accuracy and rejection accuracy, and a 0.334 abstention rate, reflecting the proportion of correctly answered questions where the answer is ``None of the choices provided'' (40 Questions / 120 Total Questions).}
    \label{fig:overall_vs_none}
\end{figure*}

\subsection{Overall Model Performance} As illustrated in Tables \ref{tab:accuracy_score} and \ref{tab:answer_score}, and Figures \ref{fig:total_score} and \ref{fig:radar_chart}, ChatGPT-o1 was of the highest performance, achieving an accuracy of 0.825, followed by Gemini 2.0 Flash Experimental (0.717) and ChatGPT-4o (0.691). The lowest-performing models were Janus 1B and Janus 7B, with accuracy scores of 0.383 and 0.433, respectively. Grok 3, despite its large parameter count of 2.7 trillion, showed disappointing results with an accuracy of 0.580, underperforming in tasks requiring complex reasoning and consistency.

Additionally, QVQ-72B-Preview demonstrated high performance in geographic understanding, but struggled with image-text matching and difference spotting, which highlights a domain-specific strength but overall inconsistency. Qwen2.5-VL-72B-Instruct displayed moderate accuracy (0.633) but encountered challenges in cartoon understanding, likely due to content restrictions preventing the processing of specific images.

The difference in accuracy between the highest- and lowest-performing models is 0.442, highlighting a substantial performance gap across evaluated models. ChatGPT-o1 leads by a margin of 13.3 percentage points over the second-best model, Gemini 2.0 Flash Experimental.

Additionally, as seen in Figure \ref{fig:radar_chart}, ChatGPT-o1 maintains the most balanced high performance across tasks, exhibiting consistently strong accuracy without significant weaknesses in any category. In contrast, models like Pixtral 12B and Grok 3 display more uneven performance, excelling in certain tasks while struggling in others. Pixtral 12B demonstrates relatively strong results in tasks such as cartoon understanding and diagram interpretation but underperforms in difference spotting and image-text matching.

\subsection{Rejection Accuracy and Abstention Rate}

As seen in Table \ref{tab:none_answer_accuracy} and Fig \ref{fig:overall_vs_none}, QVQ-72B-Preview recorded the highest rejection accuracy at 0.850, followed by ChatGPT-o1 at 0.700. Janus 7B (0.150) and Janus 1B (0.050) had the lowest rejection accuracies, showing a strong bias toward selecting an option.

For Abstention Rate, QVQ-72B-Preview had the highest rate at 0.425, followed by Grok 3 (0.375). Both exceed the threshold of 0.33, which corresponds to the proportion of questions where the correct answer was ``None of the provided options." This indicates a tendency to over-reject. In contrast, Janus 1B had the lowest abstention rate at 0.042, showing a strong reluctance to abstain even when uncertain. ChatGPT-o1 maintained a moderate abstention rate of 0.267, staying below the 0.33 threshold and demonstrating balanced uncertainty management.

\subsection{Overall Accuracy vs. Rejection Accuracy and Abstention Rate}

ChatGPT-o1 achieved both high total accuracy (0.825) and balanced rejection accuracy (0.700) with an abstention rate of 0.267, showing well-calibrated uncertainty handling. QVQ-72B-Preview, despite high rejection accuracy (0.850), had an abstention rate (0.425) above the 0.33 threshold, suggesting overrejection. Grok 3 also showed an abstention rate of 0.375, indicating a tendency to overreject despite its high parameter count.

Janus 1B displayed the most extreme behavior with the lowest rejection accuracy (0.050) and abstention rate (0.042), reflecting an overconfident approach. ChatGPT-4o and Gemini 2.0 Flash Experimental both showed moderate abstention rates below 0.33, maintaining a balanced decision strategy.

\subsection{Reasoning Stability: Entropy-Based Consistency}

To assess reasoning consistency across reordered answer variations, we compute entropy scores for each model. As shown in Table \ref{tab:miqv}, models with higher entropy scores, such as Janus 7B (0.8392), Janus 1B (0.787), and Pixtral 12B (0.557), exhibited greater response variability, suggesting fluctuations in answer selection across question variants.

In contrast, ChatGPT-o1 (0.1352), ChatGPT-4o (0.216), and Grok 3 (0.256) achieved the lowest entropy values, indicating the most consistent reasoning patterns. These results suggest that ChatGPT and $\mathbb{X}$AI \normalsize models maintain more stable reasoning, whereas other models are more influenced by positional biases or random variations in answer selection.

\section{Discussion}\label{sec:discussion}

The evaluation of multimodal language models on this dataset reveals critical insights into their reasoning capabilities, robustness, and inherent biases. This section examines whether model size correlates with performance, the effectiveness of open-source models compared to proprietary ones, the restrictive nature of Qwen models, and the impact of reordered answers in detecting biases. Additionally, we discuss the performance of Grok 3 and DeepSeek's Janus. While formal scientific validation of Grok 3 and DeepSeek's Janus's advanced capabilities is still emerging, online discussions suggest significant anticipation for their performance. Our analysis reveals that Grok 3 and DeepSeek's Janus exhibited suboptimal performance.

\subsection{Does Model Size Correlate with Performance?}

A common misconception in the research community is larger models perform better due to increased capacity for processing and learning complex relationships. While this trend holds in many cases, our results suggest that size alone does not guarantee superior performance.

In this benchmark, the smallest models, Janus 7B and Janus 1B, significantly underperformed against their larger counterparts. Given their lower parameter count, this is expected, yet their frequent positional biases, and inconsistent reasoning indicate that their training or optimization was insufficient to compensate for their size limitations.

However, Grok 3, despite having the largest parameter count of 2.7 trillion, showcased low performance relative to its size, underperforming in tasks requiring complex reasoning and consistency. Its moderate rejection accuracy (0.525) indicate inconsistent reasoning stability, suggesting that sheer model size does not necessarily lead to more robust performance or consistent decision-making.

These findings underscore that while larger models generally have greater capacity, effective optimization and fine-tuning are critical for maximizing performance and consistency.

\subsection{Are Open-Source Models Competitive Against Proprietary Models?}

While open-source models offer transparency and adaptability, this benchmark demonstrates that proprietary models continue to hold a significant advantage in complex multimodal reasoning tasks. ChatGPT-o1 and Gemini 2.0 Flash outperformed all open-source models across most reasoning tasks, underscoring the benefits of high-quality fine-tuning and diverse pretraining data. In contrast, open-source models like Janus 1B and Janus 7B exhibited significant reasoning inconsistencies (i.e., hallucination) and biases, suggesting that limited access to high-quality data remains a major constraint for non-proprietary models. However, Pixtral 12B emerged as a surprising exception, performing well in specific tasks such as Cartoon Understanding, demonstrating that with targeted fine-tuning and optimization, open-source models can still be competitive in certain domains.

\subsection{The Restrictive Nature of Qwen Models}

One of the most unexpected findings was the high rate of unanswered or incorrectly answered questions from Qwen2.5-VL-72B-Instruct and QVQ-72B-Preview due to potential content restrictions. This issue was particularly evident in Cartoon Understanding, where the model struggled with half of the test cases despite the images being non-explicit or hazardous.

This suggests that Qwen’s content filtering mechanisms are overly aggressive, preventing it from engaging with conventional content. This restrictive nature severely limits Qwen's applicability in real-world multimodal reasoning, particularly in cases where understanding humor, memes, or non-literal content is necessary. A more balanced content moderation approach could improve its usability in diverse applications without compromising ethical safeguards.

\subsection{Grok 3's Unmet Expectations} 

Grok 3, currently in its Beta version, was positioned as a powerful contender to ChatGPT-o1, boasting an impressive 2.7 trillion parameters, the largest in this benchmark. Despite its scale and claims of advanced reasoning, Grok 3 fell short of expectations, achieving underwhelming results taking its size into consideration.

Additionally, Grok 3 exhibited an unusually high abstention rate, indicating a tendency to reject answers more frequently than average, even when correct options were available. This suggests an overly conservative approach to uncertainty, which undermined its decision-making effectiveness.

While it showed moderate success in specific tasks, Grok 3 failed to outperform ChatGPT-o1 or leverage its scale for superior reasoning stability. Overall, Grok 3, although it is still in its beta version, demonstrates that model size alone does not equate to improved accuracy or reasoning consistency.

\subsection{The Underperformance of DeepSeek Janus Models}

The release of DeepSeek’s Janus models was widely anticipated, particularly following the success of DeepSeek’s R1 model, which made significant strides in competing with ChatGPT-o1 despite utilizing far fewer resources \cite{guo2025deepseek}. This led to growing expectations that DeepSeek could emerge as a serious competitor to OpenAI across multiple AI domains. However, this benchmark demonstrates that such expectations do not hold up in visual reasoning tasks. 

Janus 7B and Janus 1B, the smallest models in this benchmark, also rank among the weakest performers, struggling significantly in numerous tasks, where they exhibit severe positional biases and incosistent reasoning. The accuracy evaluation shows that Janus models fail to generalize well across answer variations, reinforcing that their reasoning ability is not robust to minor changes in answer presentation.

While some expected DeepSeek Janus to be a strong open-source alternative to ChatGPT models, its current iteration suggests that DeepSeek’s approach does not yet scale effectively to multimodal reasoning. This indicates that DeepSeek’s advancements in language modeling with R1 do not yet extend effectively to multimodal reasoning, leaving the Janus models far behind proprietary competitors like ChatGPT-o1 in this domain.

\subsection{The Continued Dominance of ChatGPT Models}

While there has been significant development in multimodal models across different organizations, ChatGPT models continue to demonstrate an overwhelming advantage in visual reasoning tasks. ChatGPT-4o and ChatGPT-o1 outperformed every other model in almost all reasoning tasks, particularly excelling in Diagram Understanding, Image-Text Matching, and Visual Retrieval. The accuracy table further highlights that these models maintained stable performance across different variations of the same questions, indicating their robustness in reasoning and comprehension.

These models consistently provided the most stable and accurate responses, exhibiting the lowest rates of hallucination and positional bias. While some open-source models, such as Pixtral 12B, showed promising results in specific areas, none were able to compete across the board with OpenAI’s ChatGPT models. This result reinforces the idea that proprietary models still maintain a significant edge in multimodal visual reasoning, likely due to access to better training datasets, fine-tuning techniques, and alignment strategies that remain unavailable to the public. Future developments in open-source multimodal models will need to focus on improving robustness and reducing bias to close the gap with proprietary alternatives.

\subsection{The Contribution of Reordered Answers and Entropy in Detecting Biases}

One of the key contributions of this study was the inclusion of reordered answer variants and the application of entropy to detect positional biases and randomness in model responses. This combined approach exposed significant differences in reasoning stability across models, highlighting the extent to which certain models rely on positional heuristics and arbitrary answers rather than genuine comprehension.

By using entropy to quantify variability in answer distributions across reordered variants, we effectively measured reasoning consistency and stability. Models with high entropy scores, such as Janus 7B (0.8392) and Janus 1B (0.787), exhibited significant inconsistencies across reordered variants, suggesting reliance on positional heuristics or randomness rather than content-based reasoning. Their fluctuations indicate that answer selection was influenced by surface-level patterns rather than a stable understanding of the question, particularly in tasks requiring multi-image integration.

In contrast, ChatGPT-o1 (0.1352) and ChatGPT-4o (0.216) demonstrated the lowest entropy scores, indicating strong reasoning stability and resistance to positional biases. These models consistently selected the same answer across reordered variations, suggesting they engage in more content-driven decision-making. While low entropy does not inherently guarantee higher accuracy, most of the top-performing models exhibited lower entropy values, showing a strong correlation between reasoning stability and overall model performance.

These findings highlight the limitations of traditional VQA evaluations, which measure correctness but not reasoning stability. By combining reordered answer variants with entropy as a consistency metric, this study provides a more nuanced and robust assessment of reasoning stability. This dual approach distinguishes models that genuinely comprehend content from those that rely on answer positioning and randomness. Future multimodal benchmarks should incorporate both reordered answers and entropy-based consistency metrics to assess reasoning robustness more effectively and mitigate heuristic-driven decision-making.

\subsection{Avoidance of ``None of the Choices Provided"}

The majority of models demonstrated a consistent tendency to avoid selecting ``None of the choices provided," even when it was the correct answer. This pattern was most pronounced in Janus 7B and Janus 1B, which exhibited extremely low rejection accuracy and abstention rates. Their strong bias towards selecting an option, regardless of correctness, suggests an overcommitment to answers, likely influenced by pretraining datasets that emphasize choosing the best available choice.

In contrast, QVQ-72B-Preview and Grok 3 displayed the highest abstention rates, exceeding the 0.33 threshold for the proportion of questions where ``None of the provided options" was correct. This indicates a more conservative approach to decision-making, with a tendency to over-reject. QVQ-72B-Preview consistently identified unanswerable questions, while Grok 3 exhibited more variability, reflecting inconsistent uncertainty calibration.

ChatGPT-o1 and ChatGPT-4o maintained balanced rejection reasoning, selectively abstaining without excessive avoidance. Their abstention rates remained close to the 0.33 threshold, indicating well-calibrated uncertainty recognition and strategic decision-making.

These findings reveal distinct patterns in rejection behavior, highlighting that while some models consistently avoid selecting ``None of the choices provided," others exhibit over-rejection tendencies. The balanced approach observed in ChatGPT models underscores the importance of effective uncertainty calibration for strategic rejection-based reasoning.

\section{Conclusion} \label{sec:conclusion}

This study provides a comprehensive evaluation of multimodal LLMs, revealing significant differences in reasoning stability, bias susceptibility, and uncertainty handling. ChatGPT-o1 and ChatGPT-4o consistently outperformed other models, demonstrating superior consistency, balanced rejection reasoning, and effective uncertainty calibration. These results highlight the advantages of extensive fine-tuning and high-quality training data in proprietary models.

Grok 3, despite its massive parameter count of 2.7 trillion, failed to meet expectations, showcasing inconsistent reasoning stability, excessive rejection behavior, and moderate overall accuracy. Its high abstention rate (0.375) indicates an overly conservative approach, emphasizing that scale alone does not guarantee better performance. Similarly, Janus 7B and Janus 1B displayed the lowest rejection accuracy and reluctance to abstain, reflecting a bias towards overcommitting to answers, likely due to insufficient exposure to rejection-based reasoning.

This study also highlights the impact of reordered answer variations in detecting positional biases. Models with high entropy scores, such as Janus 7B (0.8392) and Janus 1B (0.787), exhibited greater variability and susceptibility to positional heuristics, whereas ChatGPT-o1 (0.1352) and ChatGPT-4o (0.216) maintained consistent reasoning patterns. The introduction of entropy as a reasoning consistency metric provides a novel, quantitative measure of stability across reordered variants, revealing limitations in traditional VQA metrics that focus solely on correctness.

Rejection accuracy and abstention rates further exposed weaknesses in uncertainty calibration. QVQ-72B-Preview displayed the highest rejection accuracy but also over-rejected, exceeding the 0.33 threshold, reflecting risk-averse decision-making. Conversely, Janus models consistently avoided rejection, highlighting poor uncertainty recognition. The balanced rejection strategies of ChatGPT-o1 and ChatGPT-4o illustrate the importance of strategic abstention for reliable decision-making.

Overall, this study underscores the need for advanced benchmarks that incorporate reordered answers, entropy-based consistency metrics, and rejection accuracy to more effectively evaluate reasoning stability and uncertainty calibration. Addressing positional biases, refining rejection strategies, and enhancing generalization are crucial for advancing multimodal LLMs' real-world applicability.

{\bibliographystyle{abbrv}
\bibliography{AbdulRef}}

\end{document}